\documentclass[10pt,journal,compsoc]{IEEEtran}
\ifCLASSOPTIONcompsoc
\usepackage[nocompress]{cite}
\else
\usepackage{cite}
\fi
\ifCLASSINFOpdf
\else
\fi

\usepackage{times}
\usepackage{epsfig}
\usepackage{graphicx}
\usepackage{amsmath}
\usepackage{amssymb}
\usepackage{subfigure}
\usepackage{multirow}
\usepackage{caption}
\usepackage{url}

\begin{document}
	%
	\title{Spontaneous Facial Micro-Expression Recognition using Discriminative Spatiotemporal Local Binary Pattern with an Improved Integral Projection}
	%
	%
	%
	%
	
	\author{Xiaohua Huang, Sujing Wang, Xin Liu, Guoying Zhao, Xiaoyi Feng and Matti Pietik\"{a}inen
		\IEEEcompsocitemizethanks{\IEEEcompsocthanksitem X. Huang, X. Liu, G. Zhao and M. Pietik\"{a}inen are with the Center for Machine Vision and Signal Analysis, University of Oulu, Finland.\protect\\
			E-mail: xiaohua.huang@ee.oulu.fi, xliu@ee.oulu.fi, gyzhao@ee.oulu.fi, mkp@ee.oulu.fi
			\IEEEcompsocthanksitem S.J. Wang is with State Key Laboratory of Brain and Cognitive Science, Institute of Psychology, Chinese Academy of Science,
			Beijing, China.\protect\\
			E-mail: wangsujing@psych.ac.cn
			\IEEEcompsocthanksitem X. Feng is with School of Electronic and Information, Northwestern Polytechnic University, Xi'an, China.\protect\\
			E-mail: fengxiao@nwpu.edu.cn}
		\thanks{This works was supported by Academy of Finland, Tekes Fidipro Program and Infotech Oulu. This work was supported in part by grants from the National Natural Science Foundation of China (61379095), the Beijing Natural Science Foundation (4152055). } 
	}
	
	%
	%

	\markboth{Journal of \LaTeX\ Class Files,~Vol.~14, No.~8, August~2015}%
	{Shell \MakeLowercase{\textit{et al.}}: Bare Demo of IEEEtran.cls for Computer Society Journals}
	%



	\IEEEtitleabstractindextext{%
		\begin{abstract}
		
			Recently, there are increasing interests in inferring mirco-expression from facial image sequences. Due to subtle facial movement of micro-expressions, feature extraction has become an important and critical issue for spontaneous facial micro-expression recognition. Recent works usually used spatiotemporal local binary pattern for micro-expression analysis. However, the commonly used spatiotemporal local binary pattern considers dynamic texture information to represent face images while misses the shape attribute of face images. On the other hand, their works extracted the spatiotemporal features from the global face regions, which ignore the discriminative information between two micro-expression classes. The above-mentioned problems seriously limit the application of spatiotemporal local binary pattern on micro-expression recognition. In this paper, we propose a discriminative spatiotemporal local binary pattern based on an improved integral projection to resolve the problems of spatiotemporal local binary pattern for micro-expression recognition. Firstly, we develop an improved integral projection for preserving the shape attribute of micro-expressions. Furthermore, an improved integral projection is incorporated with local binary pattern operators across spatial and temporal domains. Specifically, we extract the novel spatiotemporal features incorporating shape attributes into spatiotemporal texture features. For increasing the discrimination of micro-expressions, we propose a new feature selection based on Laplacian method to extract the discriminative information for facial micro-expression recognition. Intensive experiments are conducted on three availably published micro-expression databases including CASME, CASME2 and SMIC databases. We compare our method with the state-of-the-art algorithms. Experimental results demonstrate that our proposed method achieves promising performance for micro-expression recognition.
			
		\end{abstract}
		
		}

		\maketitle

		\IEEEdisplaynontitleabstractindextext

		%
		\IEEEpeerreviewmaketitle

		\IEEEraisesectionheading{\section{Introduction}\label{sec:introduction}}

		Micro-expressions amongst nonverbal behavior like gestures and voice have received increasing attention in recent years~\cite{Warren}. In situations in which people are motivated to conceal or suppress their true emotions, their facial expressions may leak despite that they try to conceal them. These leakages can be very useful for true emotion analysis and many of these leakages are manifested in the form of micro-expressions. However, micro-expressions are very short involuntary facial expressions that reveal emotions people try to hide. Generally, it lasts 1/25 to 1/3 second. It is important to note that due to the visual differences of human beings, not all people reach the same level of ability to detect these facial expressions. Currently only highly trained individuals are able to distinguish them, but even with proper training the recognition accuracy is still less than 50\%~\cite{Frank}. Therefore, this poor performance makes an automatic micro-expression recognition system very attractive. 
		
		Several earlier studies on automatic facial micro-expression analysis primarily focused on distinguishing facial micro-expressions from macro-expressions~\cite{Shreve09}~\cite{Shreve11}. Shreve \textit{et al.}~\cite{Shreve09}~\cite{Shreve11} used an optical flow method for automatic micro-expression spotting on their own database. However, their database contains 100 clips of posed micro-expressions, which were obtained by asking participants to mimic some example videos that contain micro-expressions. Pollikovsky \textit{et al.} in~\cite{Polikovsky} proposed to use a 3D-gradient orientation histogram for action unit recognition on their collected database. Unfortunately, their work focused on pose micro-expression as well, since in the collection procedure they asked subjects to perform seven basic emotions with low intensity and go back to neutral expression as quickly as possible. Wu \textit{et al.} in~\cite{Wu2011} combined Gentleboost and a support vector machine classifier to recognize synthetic micro-expression samples from the Micro Expression Training Tool. The significant problem of posed micro-expressions is that they are different from real naturally occurring spontaneous micro-expressions. A study in~\cite{Ekman1969} shows that spontaneous micro-expression occurs involuntarily, and that the producers of the micro-expressions usually do not even realize that they have presented such an emotion. Therefore, methods trained on posed micro-expression cannot really solve the problem of automatic micro-expression analysis in practice.
		
		 Recently, researchers have started to conduct analysis on spontaneous facial micro-expression~\cite{Yan2013}~\cite{Li}~\cite{Pfister}~\cite{Yan2014}, as spontaneous micro-expressions can reveal genuine emotions which people try to conceal. As we know, geometry-based and appearance-based features have been commonly employed to analyze facial expressions. Recently, deep learning based methods are also proposed to recognize facial expression~\cite{Jung2015}~\cite{Yu2015}. However, geometric-based features cannot accurately capture subtle facial movements (\textit{e.g.}, the eye wrinkles) of micro-expression recognition. Deep learning based features require a large-scale database for training networks such as deep convolutional neural network. Unfortunately, the available micro-expression databases contain small number of samples, which limits the application of deep learning for micro-expression recognition. Instead, appearance-based features describe the skin texture of faces, which can capture subtle appearance changes such as wrinkles and shading changes. Amongst appearance-based features, local binary pattern (\textbf{LBP}) has been commonly used in face recognition~\cite{Ahonen2006} and facial expression recognition~\cite{Shan09} and also demonstrated its efficiency for facial expression recognition. Recently, LBP is extended to spatiotemporal domain for texture recognition and facial expression recognition~\cite{Zhao07}, which is named Local binary pattern from three orthogonal planes (\textbf{LBP-TOP}). LBP-TOP has shown its promising performance for facial expression recognition~\cite{Zhao07}~\cite{Huang2012}. Therefore, many researchers have actively focused on the potential ability of LBP-TOP for micro-expression recognition. Pfister \textit{et al.}~\cite{Pfister} proposed to use LBP-TOP for analyzing spontaneous micro-expression recognition and conducted experiments on spontaneous micro-expression corpous (\textbf{SMIC}) database. The system is the first one to automatically analyze spontaneous facial micro-expressions. It primarily consists of a temporal interpolation model and feature extraction based on LBP-TOP. In~\cite{Li}, Li \textit{et al.} continued implementing LBP-TOP on the full version of SMIC and obtained the recognition result of 48.48\%. Meanwhile, Yan \textit{et al.}~\cite{Yan2014} used the method of~\cite{Li} as the baseline algorithm on the second version of Chinese Academy of Sciences Micro-expression (\textbf{CASME2}) database. Since then, LBP and its variants have often been employed as the feature descriptors for micro-expression recognition in many other studies. For example, Davison \textit{et al.}~\cite{Davison2014} exploited  LBP-TOP to investigate whether micro-facial movement sequences can be differentiated from neutral face sequences.
		 		 
		 However, according to~\cite{Li}~\cite{Pfister}~\cite{Yan2014}  we observed that there is a gap to achieve a high-performance micro-expression analysis using LBP-TOP, since LBP-TOP has only exploited the pixel information of an image. Consequently, many works have attempted to improve the LBP-TOP. Ruiz-Hernandez and Pietik\"{a}inen~\cite{Hernandez} used the re-parameterization of second order Gaussian jet on the LBP-TOP achieving promising micro-expression recognition result on the first version of SMIC database~\cite{Pfister}. As well, Wang \textit{et al.}~\cite{Wang2014} extracted Tensor features from Tensor Independent Colour Space (\textbf{TICS}) for micro-expression recognition, but their results on the CASME2 database showed no improvement comparing with the previous results. Furthermore, Wang \textit{et al.}~\cite{Wang2014b} used Local Spatiotemporal Directional Features with robust principal component analysis for micro-expressions. Recent work in~\cite{Wang2014c} reduced redundant information in LBP-TOP by using six intersection points (\textbf{LBP-SIP}) and obtained better performance than LBP-TOP. Guo \textit{et al.}~\cite{Guo2016} employed Centralized Binary Patterns from Three Orthogonal Panels with extreme  learning machine to recognize micro-expressions. Oh \textit{et al.}~\cite{Oh2015} employed Riesz wavelet transform to obtain multi-scale monogenic wavelets for micro-expression recognition. Huang \textit{et al.}~\cite{Huang2016} proposed spatiotemporal completed local binary pattern (\textbf{STCLQP}), which utilizes magnitude and orientation as additional source and flexible encoding algorithm for improving LBP-TOP on micro-expression recognition. 
		 
		 It is noted that the most of the previous methods used in micro-expression recognition have two critical problems to be resolved. Firstly, they exploited the variance of LBP-TOP in which dynamic texture information is considered to represent face images. However, they missed the shape attribute of face images. Recent study in~\cite{Kotsia2008} suggests that the fusion of texture and shape information can perform better results than only using appearance features for facial expression recognition. Moreover, the work in~\cite{Houam} demonstrated that LBP enhanced by shape information can distinguish an image with different shape from those with the same LBP feature distributions. The method of~\cite{Houam} has been used for to achieve better performance than LBP for face recognition~\cite{Benzaoui2013}~\cite{Benzaoui2015}. Secondly, the methods in~\cite{Li}~\cite{Pfister}~\cite{Yan2014}~\cite{Huang2016} used a block-based approach for spatiotemporal features. Specifically, they firstly divide a video clip into some blocks and then concatenated features from all blocks into one feature vector. However, we observe that 
		 the dimensionality of feature may be huge. On the other hand, the same contribution from all block features would decrease the performance. Normally speaking, all spatial temporal features do not contribute equally. Therefore, in the present paper we aim to develop a new method simultaneously incorporating the shape attribute with LBP and considering the discriminative information for micro-expression recognition. 
 		 
		 Image projection techniques are classical methods for pattern analysis, widely used, \textit{e.g.}, in motion estimation~\cite{Robinson} and face tracking~\cite{Mateos03}~\cite{Mateos07}, as they enhance shape properties and increase discrimination of images. One image projection technique, integral projection, provides simple and efficient computation as well as a very interesting set of properties. It firstly is invariant to a number of image transformations~\cite{Carcia2002}. It is also highly robust to white noise~\cite{Robinson}. Then it preserves the principle of locality of pixels and sufficient information in the process of projection. Recently, integral projection is used to incorporate with LBP for achieving promising performance in bone texture classification~\cite{Houam} and face recognition~\cite{Benzaoui2013}~\cite{Benzaoui2015}. In present paper, we propose a new spatiotemporal feature descriptor, which incorporates shape attribute into dynamic texture information for improving the performance of micro-expression recognition.

		For simplicity, for extracting discriminant of feature, we may employ dimensionality reduction methods such as Linear Discriminative Analysis. However, these approaches may fail to work on micro-expression recognition because of few class number and high dimensionality of micro-expression. Zhao \textit{et al.}~\cite{Zhao2009} proposed a novel method based on AdaBoost to select the discriminative slices for facial expression recognition. But we observe that AdaBoost did not consider the closeness between two micro-expression samples and is not stable to micro-expression recognition. Recently, Laplacian method~\cite{He2005} is presented to select more compact and discriminative feature for face recognition. It considers the discriminative information and the closeness of two samples through a weighted graph. Therefore, based on the framework of~\cite{Zhao2009}, we propose a new method based on Laplacian method to learn the discriminative group-based features for micro-expression analysis.
		 		 	
		 To explain the concepts of our approach, the paper is organized as follows. In Section~\ref{Methods}, we explain our method of exploring the spatiotemporal features and discriminative information for micro-expression analysis. The results of applying our method for recognizing micro-expressions are provided in Section~\ref{experiments}. Finally we summarize our findings in Section~\ref{conclusion}.

	\section{Proposed methodology}
	\label{Methods}
	Recently, the combination of the integral projection and texture descriptor was applied to bone texture characterization~\cite{Houam} and face recognition~\cite{Benzaoui}. They demonstrate the texture descriptor is enhanced by shape information extracted by the integral projection. However, we observe that integral projection mainly represents subject information so that it cannot be directly used to describe the shape attribute of micro-expressions. In this section, we firstly resolve the problem of integral projection and then propose a new spatiotemporal feature descriptor for micro-expression recognition.
	
	\subsection{Problem Setting}
	
	\begin{figure*}[t!]
			\centering
				\includegraphics[width=\linewidth]{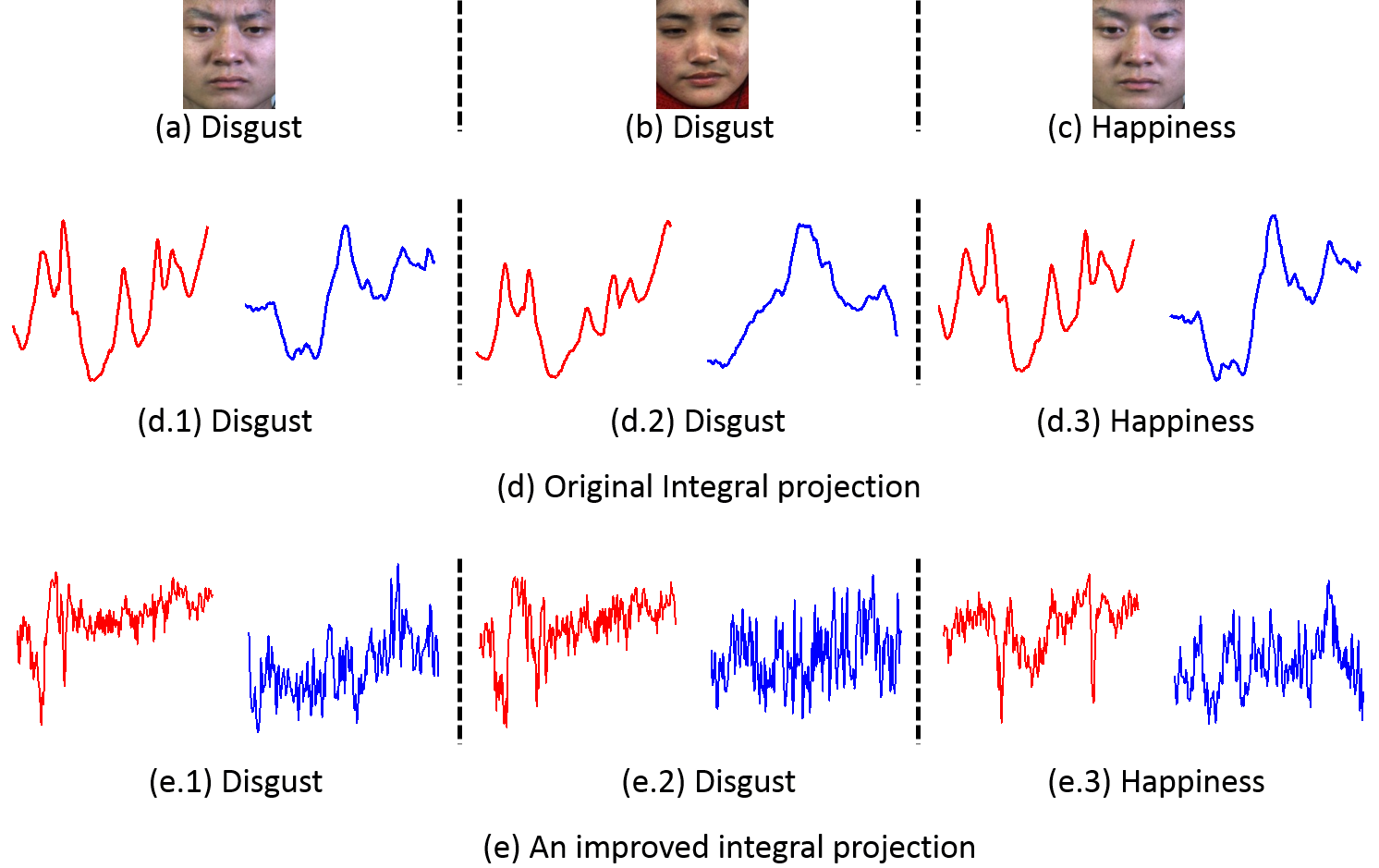}		
				\captionsetup{justification=centering}
			\caption{Integral projection and an improved integral projection on micro-expression images, where the red and blue colors represent the horizontal and vertical integral projections, respectively. (d.1) and (e.1) are the original and an improved integral projections for (a), respectively; (d.2) and (e.2) are the original and an improved integral projections for (b), respectively; (d.3) and (e.3) are the original and an improved integral projections for (c), respectively;} 
				
			\label{fig:IPDrawback}
	\end{figure*}
	
	An integral projection produces a one-dimensional pattern, obtained through the sum of a given set of pixels along a given direction. Let $I_{t}(x,y)$ be the intensity of a pixel at location $(x,y)$ and time $t$, the random transformation of $f_{t}(x,y)$ is defined as:
	\begin{equation}
	\Re[f_{t}](\theta, s)=\int_{-\infty}^{\infty}\int_{-\infty}^{\infty}I_{t}(x,y)\delta(x\cos\theta+y\sin\theta-s)dxdy,
	\label{eqn1}
	\end{equation}
	where $\delta$ is a Dirac's delta, $\theta$ is a projection angle and $s$ is the threshold value. In this work, we consider the horizontal and vertical directions. In this case, $\theta$  on Equation~\ref{eqn1} are $0^\circ$ and $90^\circ$ for horizontal and vertical directions, respectively. Thus, Equation~\ref{eqn1} can be re-written as:
		\begin{equation}
		H_t(y)=\frac{1}{x_2-x_1}\int_{x_1}^{x_2}I_{t}(x,y)dx,
		\label{eqn:HIP}
		\end{equation}
	\begin{equation}
	\label{eqn:VIP}
	V_t(x)=\frac{1}{y_2-y_1}\int_{y_1}^{y_2}I_{t}(x,y)dy,
	\end{equation}
	where $H_t(y)$ and $V_t(x)$ represent the horizontal and vertical integral projections, respectively.
	
	The pioneering work by Mateos~\textit{et al.}~\cite{Mateos03,Mateos07} said that integral projections can extract the common underlying structure for the same people's face images, which is more relative to face identity. We evaluate Equations~\ref{eqn:HIP} and~\ref{eqn:VIP} on two following cases: (1) two micro-expression images with the same class from two different persons and (2) two images with the different class from a person. They are shown in Figures~\ref{fig:IPDrawback}(d), in which the left image to the right image represent the integral projections of Figures~\ref{fig:IPDrawback}(a), Figures~\ref{fig:IPDrawback}(b) and Figures~\ref{fig:IPDrawback}(c), respectively. It demonstrates that the original integral projection cannot provide discriminative information for different micro-expressions, such as the integral projections of Figures~\ref{fig:IPDrawback}(a) and Figures~\ref{fig:IPDrawback}(c). In other words, the integral projection fails to extract the shape attribute for micro-expressions. As a result, it is necessary to improve integral projection method to obtain the class information for micro-expressions.

			\begin{figure}[t!]
				\centering
				\subfigure[]{
					\label{fig:RIP2A}
					\includegraphics[width=\linewidth]{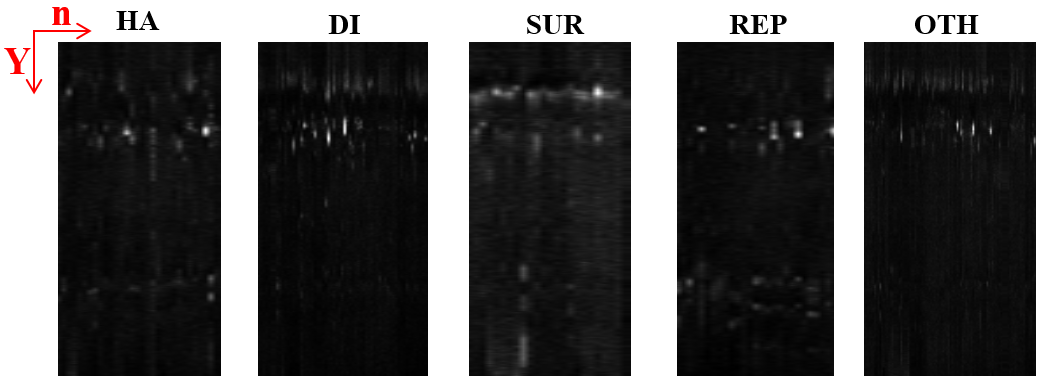}}
				\subfigure[]{
					\label{fig:RIP2B}
					\includegraphics[width=\linewidth]{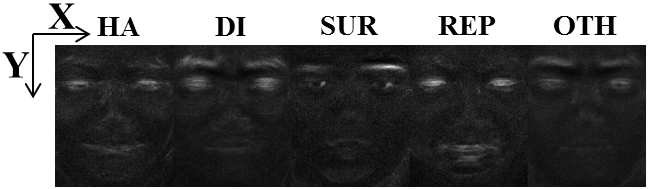}}
				\subfigure[]{
					\label{fig:RIP2C}
					\includegraphics[width=\linewidth]{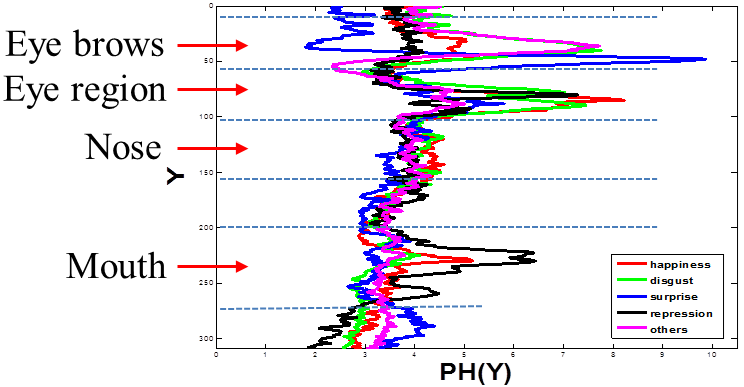}}
				\captionsetup{justification=centering}
				\caption{An improved integral projection on horizontal direction. (a) The horizontal integral projections of 247 subtle motion images, where $n$-axis describes the number of facial images and each projection is represented as a single column. (b) The mean face of the subtle motion images for happiness (HA), disgust (DI), surprise (SUR), repression (REP) and others (OTH). (c) The horizontal integral projections of the derived sparse subtle motion information in (b), where x-axis (PH(Y)) means the value of horizontal integral projection, and y-axis represents the height of an image.} 
				\label{fig:RIP2}
			\end{figure}

	\subsection{Micro-expression Augment for Integral Projection}
	\label{sec:diff}
	 For the integral projection, the key problem is how to extract the subtle facial motion information which is discriminant for recognizing micro-expression. Other information such as identity information and illumination change is not useful and instead noise for micro-expression recognition. For a micro-expression image, it can be decomposed into $Q_t(x,y)$ and $E_t(x,y)$, which is implicitly represented as following:
	\begin{equation}
	\label{eqn:MEA}
		I_t(x,y)=Q_t(x,y)+E_t(x,y),
	\end{equation}
	where $E_t(x,y)$ includes the subtle motion information of micro-expression at the $t$-th frame while $Q_t(x,y)$ is the other information. In this section, we aim to compute the integral projection on $E$ in order to resolve the problem of integral projection.
	
	For a micro-expression video clip, other information including illumination, pose and subject identity accounts for the great proportion of the whole information in a clip, while the subtle facial motion information is sparse. Equation~\ref{eqn:MEA} can be viewed to extract the sparse information for a video clip. Recently, Robust principal component analysis (RPCA) is widely used for face recognition and video frame interpolation. RPCA leverages on the fact that the data are characterized by low-rank subspaces. Therefore, we aim to extract this sparse information $E$ using RPCA~\cite{Wright2009} for the integral projection. 
	
	Given a micro-expression video clip, each of its frame is vectorized as a column of matrix $I\in\Re^{D}$. As $E$ includes the derived sparse subtle motion information, the optimization problem of Equation~\ref{eqn:MEA} is formulated as follows:
	\begin{equation}
	\label{eqn:rank}
		[Q, E]=\min\text{rank}(Q)+\parallel E\parallel_{0}, \text{w.r.t.}~~I=Q+E,
	\end{equation}
	where $\text{rank}(\cdot)$ denotes the rank of matrix and $\parallel \cdot\parallel_{0}$ means $L_0$ norm. Because of not-convex problem of Equation~\ref{eqn:rank}, it is converted into the convex optimization problem as followed: 
		\begin{equation}
		\label{eqn:RPCA}
		[Q, E]=\min\parallel S\parallel_*+\lambda\parallel E\parallel_{0}, \text{w.r.t.}~~I=Q+E,
		\end{equation}
	where $\parallel \cdot\parallel_*$ denotes the nuclear norm, which is the sum of its singular values. $\lambda$ is a positive weighting parameter.
	
	For solving Equation~\ref{eqn:RPCA}, the iterative thresholding technique can be used to minimize a combination of both the $L_0$ norm and the nuclear norm, while this scheme converges extremely slowly. Instead, Augmented Lagrange Multipliers (ALM) is more efficient way to solve Equation~\ref{eqn:RPCA}. Specifically, ALM is introduced for solving the following constrained optimization problem:
	\begin{equation}
	\label{eqn:ALM}
	X=\min f(X), \text{w.r.t.}~~h(X)=0,
	\end{equation}
	where $f:\Re^n\rightarrow\Re$ and $h:\Re^n\rightarrow\Re^m$. The augmented Lagrangian function can be defined as follows:
	\begin{equation}
	\label{eqn:LAG}
	L(X,Y,\mu)=f(X)+\langle Y,h(X)\rangle+\frac{\mu}{2}\parallel h(x)\parallel_{F}^2.
	\end{equation}
	
	Let $X$ be $(Q,E)$, $f(X)$ be $\parallel Q\parallel_*+\lambda\parallel E\parallel_{1}$, and $h(X)$ be $I-Q-E$. Equation~\ref{eqn:LAG} is re-written as followed:	
		\begin{equation}
		\begin{split}
		\label{eqn:sol}
		L(Q,E,Y,\mu) &=\parallel Q\parallel_*+\lambda\parallel E\parallel_{1} \\
		&+\langle Y,I-Q-E\rangle+\frac{\mu}{2}\parallel I-Q-E\parallel_{F}^2,
		\end{split}
		\end{equation}
	which leads two algorithm for ALM: exact ALM or inexact ALM. A slight improvement over the exact ALM leads to the inexact ALM, which converges practically as fast as the exact ALM, but the required number of partial SVDs is significantly less. In this paper, we choose inexact ALM to extract the subtle facial motion information. For integral projection, Equations~\ref{eqn:HIP} and~\ref{eqn:VIP} are re-written as
	\begin{equation}
	H_t(y)=\frac{1}{x_2-x_1}\int_{x_1}^{x_2}E_t(x,y)dx.
	\label{eqn6}
	\end{equation}
	\begin{equation}
	V_t(x)=\frac{1}{y_2-y_1}\int_{y_1}^{y_2}E_t(x,y)dy.
	\end{equation}
		
	We show the improved integral projection on three micro-expression images in Figure~\ref{fig:IPDrawback}(e). We see that two images with the same class have similar face shape along horizontal and vertical integral projections while ones with different class have distinct face shape. We also investigate the influence of RPCA to horizontal integral projection on 247 facial images of 5-class micro-expression from CASME2~\cite{Yan2014} in Figure~\ref{fig:RIP2}. In Figure~\ref{fig:RIP2A}, the horizontal integral projections from 247 images capture the various structure of signals for different micro-expressions. Additionally, they obtain the specific structure from such regions of interest of micro-expression as mouth region for happiness expression. Moreover, as seen from Figure~\ref{fig:RIP2B}, the subtle motion image obtained by RPCA well characterizes the specific regions of facial movements for different micro-expressions. For example, disgust expression mostly appears in eyebrows and eyes. Another finding in Figure~\ref{fig:RIP2C} argues the improved integral projection can preserve the discriminative structure of 1D signals for different micro-expressions. From these observations, the improved integral projection can provide more discriminative information for micro-expressions.

	\subsection{Spatiotemporal Local Binary Pattern based on Improved Integral Projection}
	\label{sec:RSLBP}
		
 	The improved integral projection preserves the shape attribute of different micro-expressions and has discriminative ability. But it is not robust to describe the appearance and motion of facial images. As LBP-TOP~\cite{Zhao07} considers micro-expression video clips from three orthogonal planes, representing appearance and motion information, respectively. We exploit the nature of LBP-TOP to obtain the appearance and motion features from the integral projections.
	\begin{figure}
		\centering
		\includegraphics[width=0.8\linewidth]{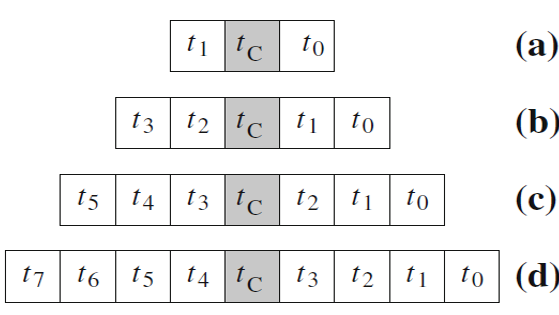}
		\captionsetup{justification=centering}
		\caption{Linearly symmetric neighbour sets for different values of $W$: (a) $W=3$; (b) $W=5$; (c) $W=7$; (d) $W=9$, where $t_c$ is the value of the center pixel.}
		\label{fig:mask}
		\vspace{-10pt}
	\end{figure}
	
	Firstly, we look at the method to extract the appearance for micro-expressions. Let $S_t$ be the integral projection at time $t$, where it can be $V_t$ or $H_t$. The appearance information of an image can be extracted by using one-dimensional local binary pattern (\textbf{1DLBP})~\cite{Houam}. The linear mask of size $W$, which can be designed as 3, 5, 7 or 9, is used to scan $S_t$ with one element step. The mask pattern is exampled in Figure~\ref{fig:mask}. The 1DLBP code is calculated by thresholding the neighborhood values against the central element. The neighbors will be assigned the value 1 if they are greater than or equal to the current element and 0 otherwise. Then each binary element of the resulting vector is multiplied by a weight depending on its position. This can be summarized as
	\begin{equation}
	\text{1DLBP}_{t,W}=\sum_{p}\delta(S_t(z_p)-S_t(z_c))2^p,
	\end{equation}
	where $\delta$ is a Dirac's delta, $S_t(z_c)$ is the value at the center of the mask, $z_c\in[y_1,y_2]$ or $[x_1,x_2]$, and $z_p$ is the neighbors of $z_c$. 
	
	\subsubsection{Spatial domain}
\begin{figure*}[t!]
		\centering
		\includegraphics[width=\linewidth]{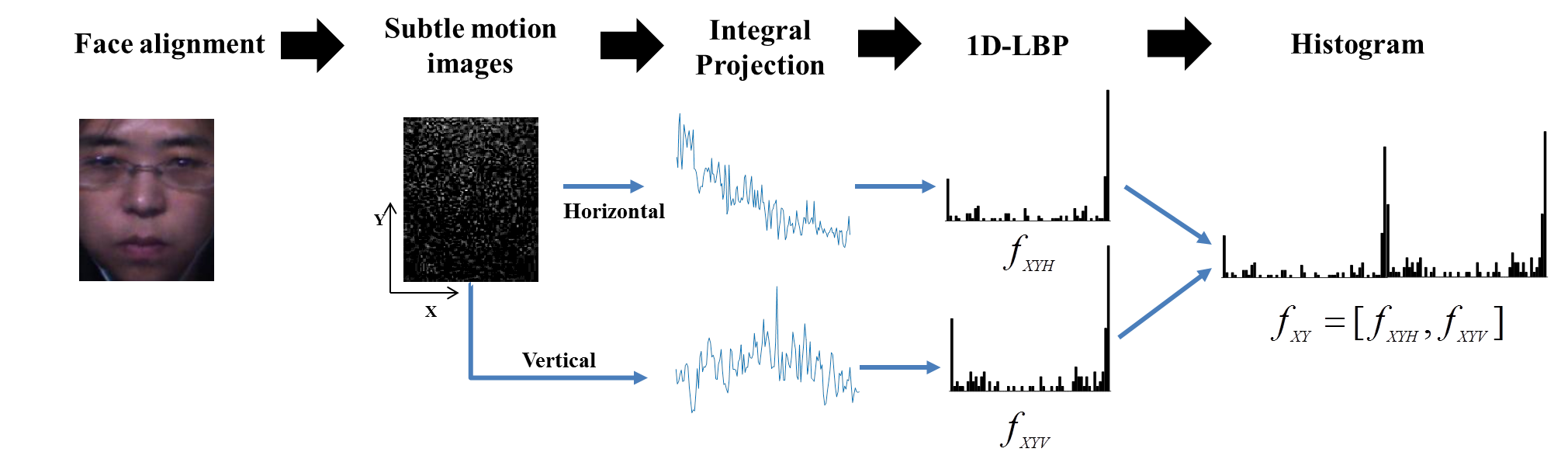}
		\captionsetup{justification=centering}
		\caption{Procedure of encoding an improved integral projection on spatial domain.}
		\label{fig:1DLBP}
	\end{figure*}

	The distribution of the 1D patterns of each frame is modeled by a histogram which characterizes the frequency of each pattern in the 1D projected signal. This encodes the local and global texture information since the projected signal handles both cues. Figure~\ref{fig:1DLBP} shows the procedure to encode the integral projection by using LBP. To describe the appearance of each video clip, the histograms of frames are accumulated. Finally the spatial histogram is represented by histogram $f_{XYH}$ and $f_{XYV}$ of horizontal and vertical projections.

	\begin{figure*}[t!]
		\centering
		\includegraphics[width=\linewidth]{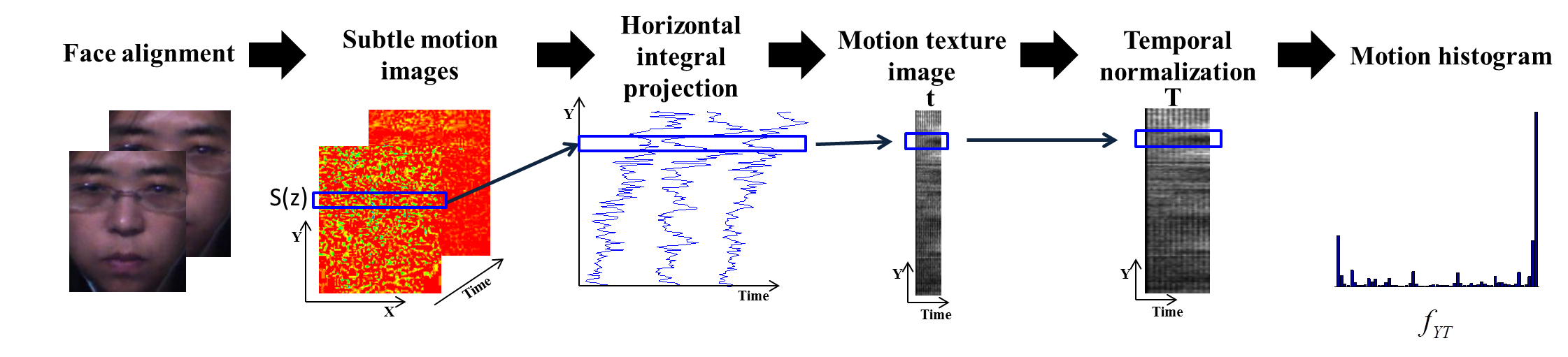}
		\captionsetup{justification=centering}
		\caption{Motion histogram along YT plane: The blue rectangle represents horizontal integral projection at specific range of Y-axis direction, where $t$ and $T$ are the original and temporal normalization time lengths, respectively.}
		\label{fig:MM}
	\end{figure*}
	
	\subsubsection{Temporal domain}
	Motion features are extracted from horizontal and vertical direction. Firstly, we consider a simple way to extract the motion histogram along horizontal direction, as shown in Figure~\ref{fig:MM}. We formulate the horizontal integral projections from all difference images in a video clip as a new texture image, which is similar to the YT plane of LBP-TOP. It represents the motion of micro-expression video clip along vertical direction. As seen from Figure~\ref{fig:MM}, the change of value $S(z)$ ($z\in[y_1,y_2]$) along the time $t$ definitely shows the motion change of shape of micro-expressions along the vertical direction. 
	However, the changing rate of micro-expression video clips might be different; it might cause unfair comparison among the motion histograms. Bilinear interpolation is utilized to ensure $S(z)$ along the time $t$ with the same size $T$. Here we name this procedure \textit{temporal normalization}. Based on the new texture image, a gray-scale invariant texture descriptor, LBP operator~\cite{Ojala02}, which is defined as
	\begin{equation}
	\text{LBP}_{M,R}=\sum_{m=0}^{M-1}\delta(g_m-g_c)2^m,
	\end{equation}
	is exploited to extract the motion histogram, where $g_c$ is the gray value of the center pixel, $g_m$ is the gray value of $M$ equally spaced pixels on a circle of radius $R$ at this center pixel. The same procedure is applied to vertical integral projection.
	
	Empirical experiments tell us that the procedure normalizing all images into the same size could produce the promising performance. It also allows us to use the same value of $R$ for motion texture images. So far, the motion histograms, which represent motion change along the vertical (YT) and horizontal (XT) directions, are obtained by the process described above. Here, we denote them as $f_{YT}$ and $f_{XT}$. 
	
	The final feature vector of a micro-expression video clip can be formulated by $[f_{XYH}$, $f_{XYV}$, $f_{XT}$, $f_{YT}]$, where this feature preserves shape and texture information. For convenience, we abbreviate Spatiotemporal local binary pattern with improved integral projection as~\textbf{STLBP-IIP}.
	
	\subsection{Enhancing discriminative ability}
	\label{sec:diSTLBP}
	
	In general, we divide micro-expression video clip into $m\times n$ blocks in spatial domain. In the $k$-th block, the feature on spatial domain contains two sub-features $f_k^{XYH}$ and $f_k^{XYV}$ obtained from horizontal $XYH$ and vertical $XYV$ directions of $XY$ plane, respectively, while the feature on temporal domain consists of $f_k^{XT}$ and $f_k^{YT}$ extracted from $XT$ and $YT$ planes, respectively. In practice, we concatenates these features into one feature vector for micro-expression recognition. However, all features do not contain equally discriminative information for different micro-expressions. For simplicity, for extracting discriminative of features, we may employ dimensionality reduction methods such as Linear Discriminative Analysis. However, these approaches may fail to work on micro-expression recognition due to few class number and high dimensionality of micro-expression features. Instead, we aim to extract the discriminative features from $\{f_k^{XYH},f_k^{XYV},f_k^{XT},f_k^{YT}\}_{k=1}^{m\times n}$ for micro-expression recognition. For convenience, we define one sub-feature from $XYH$, $XYV$, $XT$ or $YT$ as a group feature. In our method, we propose a group feature selection on the basis of Laplacian method~\cite{He2005} and pairwise-class micro-expression to extract the discriminative information of STLBP-IIP for micro-expression recognition, since Laplacian method is based on the following observation: two data points are probably related to the same class if they are close to each other. Our method consists of two important steps: (1) formulation of dissimilarity feature and (2) computation of Laplacian scores of group features.
		
	\textbf{Formulation of dissimilarity feature}: Given a micro-expression video clip, we divide it into $m\times n$ blocks in spatial domain. For the $k$-th block, its feature is represented by $f_k=[f_k^{XYH}$, $f_k^{XYV}$, $f_k^{XT}$, $f_k^{YT}]$. Thus the dissimilarity of the $i$-th and $j$-th micro-expression video clips $F_i$ and $F_j$ on the $k$-th block contains the difference between group features, which is defined as followed,
	\begin{equation}
		d_k(F_i,F_j)=[d_k^{XYH}~d_k^{XYV}~ d_k^{XT}~ d_k^{YT}],
		\label{eqn:X2}
	\end{equation}
	where $d_k^{XYH}=\chi^2(f_{i,k}^{XYH},f_{j,k}^{XYH})$, $d_k^{XYV}=\chi^2(f_{i,k}^{XYV},f_{j,k}^{XYV})$, $d_k^{XT}=\chi^2(f_{i,k}^{XT},f_{j,k}^{XT})$, $d_k^{YT}=\chi^2(f_{i,k}^{YT},f_{j,k}^{YT})$ and $\chi^2$ is Chi-square distance metric.

	Based on Equation~\ref{eqn:X2}, the new feature from any micro-expression-pair samples $g\in\Re^{m\times n\times 4}$  is formulated as $[d_1^{XYH},d_1^{XYV},d_1^{XT},d_1^{YT},\ldots,d_{m\times n}^{XYH},d_{m\times n}^{XYV},d_{m\times n}^{XT},d_{m\times n}^{YT}]$, in which one dimension describes the dissimilarity of each group feature of two different samples. Its corresponding class information for $g$ is labeled as followed,  	
	  \begin{equation}
	      \label{eqn:class}
	  c(g) = \left\{
	  \begin{array}{ll}
	 1   & \quad \text{if } c(F_i)= c(F_j)\\
	   -1  & \quad \text{if } c(F_i)\neq c(F_j),
	  \end{array}
	  \right.
	  \end{equation}
	where $c(F_i)$ and $c(F_j)$ are the class label of two micro-expression video clips $F_i$ and $F_j$.  
    
    \textbf{Computation of Laplacian scores of group features}: We employ one-vs-one class learning strategy to obtain the discriminative group feature for two micro-expression classes. Given samples from the $a$-th and $b$-th micro-expression classes, we can formulate the dissimilarity features $G=[g_1,\ldots,g_N]$ and their corresponding labels $C=[c_1,\ldots,c_N]$, where $N$ is the number of dissimilarity features, $N=N_a(N_a-1)+N_bN_b$, $N_a$ and $N_b$ are the number of samples with the $a$-th and $b$-th micro-expression classes, respectively. We construct a weighted graph $\mathcal{G}$ with edges connecting nearby points to each other, in which $W_{uv}$ evaluates the similarity between the $u$-th and $v$-th samples. In our method, we employ the class label and Cosine metric for constructing the weight matrix $W$, which models the local structure of the data space. The weight matrix $W$ is defined as:
    \begin{equation}
    	W = \left\{
    	\begin{array}{ll}
    	\frac{g_u\cdot g_v}{\parallel g_u\parallel \parallel g_v\parallel} & \quad \text{if } c(g_u) = c(g_v) \\
    	0 & \quad \text{otherwise},
    	\end{array}
    	\right.
    \end{equation}    
    where $\cdot$ is a dot product, $g_u$ and $g_v$ are the $u$-th and $v$-th samples in $G$, respectively. 
    A reasonable criterion for choosing a good feature is to minimize the following object function:
    \begin{equation}
    	L_r=\frac{\sum_{uv}(g_{ru}-g_{rv})^2W_{uv}}{Var(\mathbf{g_r})},
    \end{equation}
    where $r\in \{1,\ldots,m\times n\times 4\}$ is dimension index of feature $g$, $\mathbf{g_r}=[g_{r}^1,g_{r}^2,\ldots,g_{r}^{N}]$, and $Var(\mathbf{g_r})$ is the estimated variance of the $r$-th feature.
    
    For a good feature, the bigger $W_{uv}$, the smaller $(g_{ru}-g_{rv})$. As well, by maximizing $Var(\mathbf{g_r})$, we prefer those features with large variance which have more representative power. Thus the Laplacian Score tends to be small. According to~\cite{He2005}, $\sum_{uv}(g_{ru}-g_{rv})^2W_{uv}$ is written as
    \begin{equation}
    	\sum_{uv}(g_{ru}-g_{rv})^2W_{uv}=2\mathbf{g_r}^TD\mathbf{g_r}-2\mathbf{g_r}^TW\mathbf{g_r},
    \end{equation}
    where $D=diag(W\mathbf{1})$, $\mathbf{1}=[1,\ldots,1]^T$ and $W$ is the weight matrix containing $W_{uv}$. $Var(\mathbf{g_r})$ can be estimated as follows:
    \begin{equation}
    \label{eqn:var}
    \begin{split}
    	Var(\mathbf{g_r}) & =\sum_{u}(g_{ru}-\mu_r)^2D_{uu}\\
    	&= \sum_{u}(g_{ru}-\frac{\mathbf{g_r}^TD\mathbf{1}}{\mathbf{1}^TD\mathbf{1}})^2D_{uu}, 
    \end{split}
    \end{equation}
    after removing the mean from the samples, Equation~\ref{eqn:var} is rewritten as:
    \begin{equation}
    \label{eqn:var2}
        Var(\mathbf{g_r})=\tilde{\mathbf{g_r}}^TD\tilde{\mathbf{g_r}}.
    \end{equation}
    
    For each feature, its Laplacian score is computed to reflect its locality preserving power. Therefore, the Laplacian Score of the $r$-th feature as follows:
    \begin{equation}
    \label{eqn:obj}
    	L_r=\frac{\tilde{\mathbf{g_r}}^TL\tilde{\mathbf{g_r}}}{\tilde{\mathbf{g_r}}^TD\tilde{\mathbf{g_r}}}.
    \end{equation}
    
    Based on Equation~\ref{eqn:obj}, the Laplacian score of each group feature is calculated. The group feature with the smallest score have the strongest discriminative ability. We sort them in ascending order, and then choose the first $P$ group features for pairwise micro-expression classes. In this subsection, the group selected features of STLBP-II are named as \textbf{DiSTLBP-IIP}, respectively.

	\section{Experiments}
	\label{experiments}
	
	In this paper, we develop discriminative spatiotemporal local binary pattern based on an improved integral projection (STLBP-IIP and DiSTLBP-IIP). In this section, we evaluate STLBP-IIP and DiSTLBP-IIP on the Chinese Academy of Sciences Micro-expression Database (CASME)~\cite{Yan2013}, CASME2~\cite{Yan2014} and Spontaneous Micro-expression Corpus (SMIC)~\cite{Li}. 
		
		\begin{table*}[t!]
			\small 
			\captionsetup{justification=centering}
			\caption{Effects of the mask size $W$ to STLBP-IIP on CASME, CASME2 and SMIC databases, where the bold number means the best recognition rate (\%).}
			\begin{tabular}{|c|c|c|c|c|c|c|c|c|c|c|c|c|}
				\hline \multirow{3}{*}{Block Number}& \multicolumn{4}{|c|}{CASME} & \multicolumn{4}{|c|}{CASME2} & \multicolumn{4}{|c|}{SMIC} \\ 
				\cline{2-13}
				& \multicolumn{4}{|c|}{Mask size} & \multicolumn{4}{|c|}{Mask size} & \multicolumn{4}{|c|}{Mask size}\\
				\cline{2-13}
				& 3 & 5 & 7 & 9 & 3 & 5 & 7 & 9 & 3 & 5 & 7 & 9\\ \hline
				$6\times 1$ & 52.63 & 51.46 & 52.05 & 53.22 & 59.11 & 58.70 & 61.94 & \textbf{62.75} & 49.39 & 46.95 & 49.39 & 48.78\\ \hline
				$7\times 3$ & 54.97 & 54.39 & 55.56 & \textbf{56.14} & 53.44 & 51.82 & 53.04 & 52.23 & 49.39 & 53.05 & 47.56 & 54.88\\ \hline
				$5\times 5$ & 52.05 & 53.22 & 50.88 & 47.95 & 51.82 & 48.58 & 49.39 & 51.01  & 50.00 & 50.61 & 55.49 & 52.44\\ \hline
				$5\times 6$ & 54.39 & 54.39 & 50.88 & 52.05 & 54.66 &52.23  &52.23 & 54.25 & 57.94 & 56.10 & \textbf{59.76} & 58.54\\ \hline
				$8\times 8$ & 43.27 & 42.69 & 42.11 & 43.86 & 51.42 & 51.42 & 43.93 & 50.20 & 49.39 & 50.61 & 52.44 & 52.44\\ \hline
			\end{tabular}
			\label{tab:W4STLBPSIP}					
			\centering
		\end{table*}
	
	\subsection{Database description and protocol}
			
	The CASME dataset contains spontaneous 1,500 facial movements filmed with 60 fps camera. Among them, 195 micro-expressions were coded so that the first, peak and last frames were tagged. Referring to the work of~\cite{Yan2013}, we select 171 facial micro-expression videos that contain disgust (44 samples), surprise (20 samples), repression (38 samples) and tense (69 samples) micro-expressions.

	The CASME2 database includes 247 spontaneous facial micro-expressions recorded by a 200 fps camera and spatial resolution with $640\times 480$ pixel size. In this database, they elicited participants' facial expressions in a well-controlled laboratory environment with proper illumination. These samples are coded with the onset and offset frames, as well as tagged with AUs and emotion. There are 5 classes of the micro-expressions in this database: happiness (32 samples), surprise (25 samples), disgust (64 samples), repression (27 samples) and others (99 samples).
	
	The SMIC database consists of 16 subjects with 164 spontaneous micro-expressions recorded in a controlled scenario by 100 fps camera with resolution of 640$\times$480. 164 spontaneous facial micro-expressions are categorized into positive (51 samples), negative (70 samples) and surprise (43 samples) classes. 
	
	For three databases, we firstly use Active Shape Model to obtain the 68 facial landmarks for a facial image, and align it to a canonical frame. For the CASME and CASME2 databases, the face images are cropped to 308$\times$257 pixel size, while for the SMIC database, we crop facial images into 170$\times$139. In the experiments, we use leave-one-subject-out cross validation protocol, where the samples from one subject are used for testing, the rest for training. We use the Support Vector Machine (SVM) with Chi-Square Kernel~\cite{CC01a}, where the optimal penalty parameter is provided using the three-fold cross validation approach.

	\subsection{Evaluation of the improved integral projection}
		
		In this scenario, we compare STLBP-IIP with the previous method based on original integral projection (STLBP-OIP) and STLBP-IP~\cite{Huang2015} on CASME, CASME2 and SMIC databases. For three databases, we conducted a comparison on $7\times 3$ spatial blocks of micro-expression video clip. We set $W$ as 9 and do not use temporal normalization. All comparisons are evaluated using recognition rate.
		
		(1) On CASME database, we list the recognition rate for three spatiotemporal features, where the accuracies of 35.67\%, 54.39\% and 56.14\% for STLBP-OIP, STLBP-IP and STLBP-IIP, respectively. Comparing with STLBP-OIP and STLBP-IP, STLBP-IIP improves the performance by increasing substantially recognition rate of 18.72\% and 1.75\%, respectively. 
		
		(2) On CASME2 database, STLBP-OIP and STLBP-IP obtain the accuracy of 42.51\% and 52.63\%, respectively, while STLBP-IIP achieves the recognition rate of 56.68\%. It is seen that comparing with STLBP-OIP and STLBP-IP, STLBP-IIP increases the recognition rate of 10.12\% and 4.05\%, respectively. 
		
		(3) On SMIC database, we obtain the accuracies of 34.15\%, 45.73\% and 54.88\% for STLBP-OIP, STLBP-IP and STLBP-IIP, respectively. Comparing with STLBP-OIP and STLBP-IP, the performance of micro-expression recognition is significantly increased at the recognition rate of 11.58\% and 9.15\% by using STLBP-IIP, respectively. 
		
		It is noted that the performance is substantially improved by spatiotemporal local binary pattern using an improved integral projection method comparing with the previous method using the original integral projection. It can better reduce the influence of subject information in integral projection. The discriminative ability of integral projection can be enhanced by our proposed method. Additionally, we see that STLBP-IIP outperforms STLBP-IP on three databases, as STLBP-IIP uses robust principal component analysis to obtain more stable motion information than STLBP-IP.
	
	\subsection{Parameter evaluation}
	
	The mask size $W$ of 1DLBP, the radius $R$ of LBP and the temporal normalization size $T$ are three important parameters for STLBP-IIP, which determine the computational complexity and classification performance. Additionally, for DiSTLBP-IIP the number of selected group features $P$ decides their performance. In this subsection, we evaluate the effects of $W$, $R$, $T$ and $P$.

	\textbf{The mask size:} We evaluate the performance of STLBP-IIP caused by various $W$ on CASME, CASME2 and SMIC databases. In order to avoid bias, and to compare the performance of features on a more general level, spatiotemporal features are extracted by varying block number. It is noted that $W$ relatively controls the feature extracted on spatial domain. So temporal normalization is not considered in comparing the performance of various $W$. The results of STLBP-IIP on three databases are presented in Table~\ref{tab:W4STLBPSIP}. It is found that the performance is boosted with increasing $W$ when we use small block number for three databases. It is explained by that using more neighboring number can provide compact and much information for robust binary pattern. But for large block number, the big $W$ decreases the performance, since the feature will be very sparse due to less sampling points along horizontal/vertical integral projection for 1DLBP. 
	
	As seen from Tables~\ref{tab:W4STLBPSIP},  for STLBP-IIP, $7\times 3$, $6\times 1$ and $5\times 6$ are the optimal block number on CASME, CASME2 and SMIC databases, respectively. STLBP-IIP obtains the promising results under $W=9, 9, 7$ on CASME, CASME2 and SMIC databases, respectively.
	
	\textbf{The radius of LBP:} Based on the designed $W$ and block numbers, we evaluate the effect of $R\in\{1,2,3\}$ on CASME, CASME2 and SMIC databases. Results are presented in Table~\ref{tab:R4STLBP}. It is found that STLBP-IIP obtains the best recognition when $R=3$. 
		
					\begin{table}[t!]
						\small 
						\captionsetup{justification=centering}
						\caption{Performance of STLBP-IIP using different radius of LBP $R$ on CASME, CASME2 and SMIC databases.}
						\begin{tabular}{|c|c|c|c|}
							\hline \multirow{2}{*}{Database}& \multicolumn{3}{|c|}{$R$}\\ 
							\cline{2-4}
							& 1 & 2 & 3\\ \hline
							CASME & 50.88 & 52.05 & 56.14\\ \hline
							CASME2 & 58.70 & 59.92 & 62.75\\ \hline
							SMIC & 50.00 & 53.05 & 59.76\\ \hline
						\end{tabular}
						\label{tab:R4STLBP}					
						\centering
					\end{table}
		
\begin{figure}[t!]
	    		\centering
	    		\includegraphics[width=\linewidth]{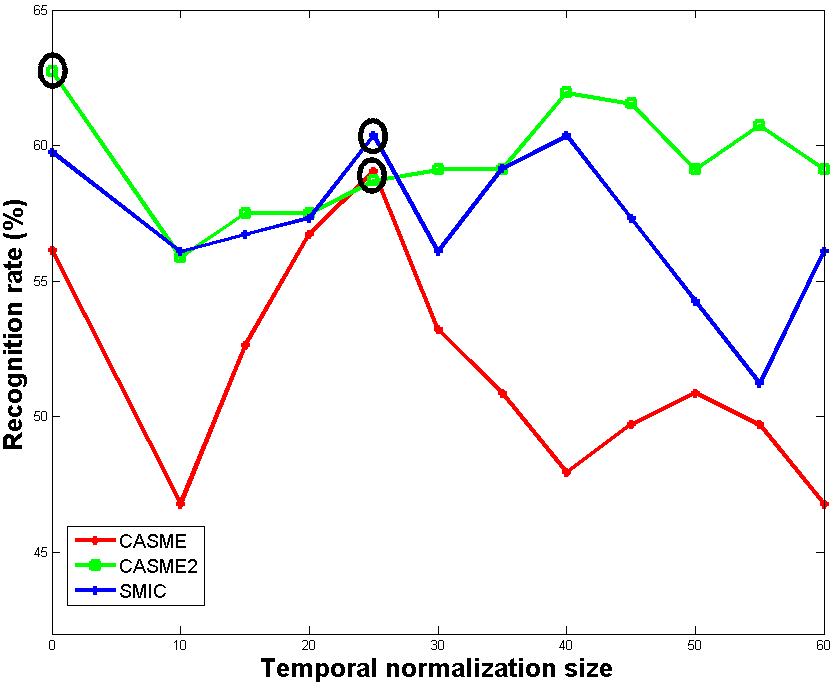}
	    		\captionsetup{justification=centering}
	    		\caption{Influence evaluation of temporal normalization size to STLBP-IIP, where the black circle means the best result.} 
	    		\label{fig:TIM4CS}	
	    		
	    	\end{figure}

	\textbf{The temporal normalization size:} Based on the well-designed $W$, $R$ and block number, we evaluate the influence of $T$ to STLBP-IIP on CASME, CASME2 and SMIC databases. All experiments are conducted under $T\in [0, 60]$, where $T=0$ means no temporal normalization used for STLBP-IIP. Figure~\ref{fig:TIM4CS} shows the effect of $T$ to STLBP-IIP on CASME, CASME2 and SMIC databases. It is seen that temporal normalization method boosts the ability of STLBP-IIP on CASME and SMIC databases, while it cannot be helpful to CASME2 database. It may be explained by that the micro-expression video clip on CASME2 database is recorded by using 200 fps camera, which provides enough temporal information. As examined from Figure~\ref{fig:TIM4CS}, STLBP-IIP obtains the highest performance on CASME and SMIC databases by using $T=25$.   
	    
		\begin{figure}[t!]
			\centering			
				\includegraphics[width=\linewidth]{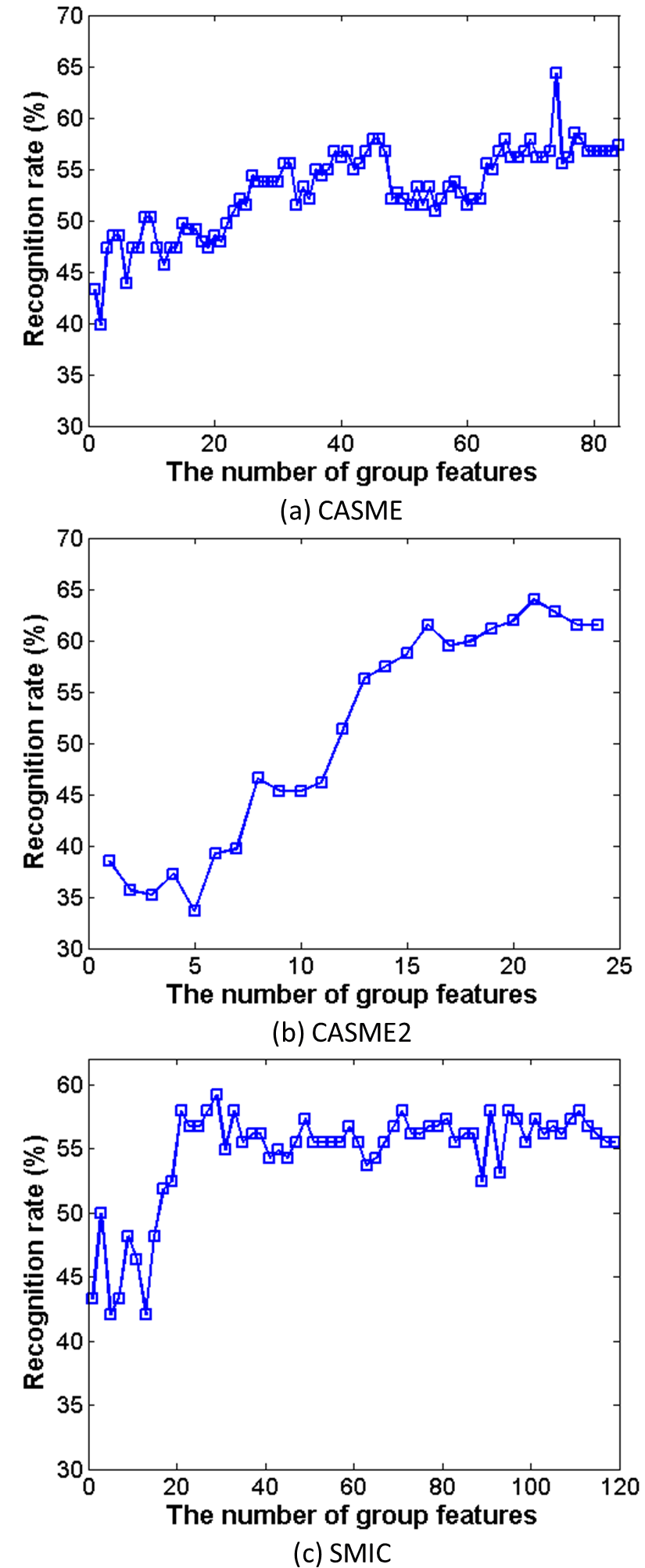}	
					\captionsetup{justification=centering}
			\caption{The influence of the number of group features to STLBP-IIP for (a) CASME, (b) CASME2 and (c) SMIC databases.} 
			\label{fig:LS}
		\end{figure}
	
    \textbf{The number of group features:} We evaluate the Laplacian method to STLBP-IIP on CASME, CASME2 and SMIC databases. The effect of the number of group features on three database is presented in Figure~\ref{fig:LS}.
	
	(1) CASME: For two features, 84  ($7\times 3\times 4$) group features are available for feature selection as micro-expression video clip is divided into $7\times 3$ across spatial domain. As shown in Figure~\ref{fig:LS}(a), the performance of STLBP-IIP is substantially improved with increasing $P$. It means that more group features can provide discriminative information, but no need to include all group features. It is noted that DiSTLBP-IIP achieves 64.33\% with 74 group features, respectively. The performance is improved at increased recognition rate of 3.04\% comparing with STLBP-IIP.
	
	(2) CASME2: Since we divide facial images into $6 \times 1$ blocks in spatial domain, we have 24 group features for STLBP-IIP. From Figure~\ref{fig:LS}(b), we can see that the recognition rate is increased with large group features number. DiSTLBP-IIP achieves 64.78\% with 21 group features. Comparing with STLBP-IIP, the performance is improved at increased recognition rate of 2.03\%, respectively. 
	
	(3) SMIC: Since we divide facial image into $5\times 6$ for STLBP-IIP, there exists 120 group features. As seen in Figure~\ref{fig:LS}(c), DiSTLBP-IIP achieves 63.41\% with 30 group features. Comparing with STLBP-IIP, the performance is improved at increased recognition rate of 3.04\%. 
	
	It shows that Laplacian method can enhance the discriminative ability of STLBP-IIP. Moreover, with promising group features, the computational efficiency becomes better, because Laplacian method reduces the dimensionality of spatiotemporal features. 
	
	We compare Laplacian method with two feature learning strategies~\cite{Zhao2009} using Boosting algorithm and Fisher linear discriminant, which are denoted as Boosting and Fisher, respectively. Boosting approach obtains 54.97\%, 53.85\% and 53.05\% for CASME, CASME2 and SMIC databases, respectively, while Fisher method achieves the recognition rate of 59.06\%, 61.94\% and 55.49\% for CASME, CASME2 and SMIC databases, respectively.
	
	We observe that Boosting algorithm failed to work for STLBP-IIP on three databases. Instead, Fisher method substantially improves STLBP-IIP. However, its performance is worse than Laplacian method. Comparisons demonstrate that Laplacian method can enhance the discriminative ability of spatiotemporal feature descriptor better than two feature selection methods in~\cite{Zhao2009}.

	\subsection{Algorithm Comparison}
	
	In this subsection, we compare STLBP-IIP and DiSLBP-IIP with the state-of-the-art algorithms on CASME, CASME2 and SMIC databases. 
	
%
		
	\subsubsection{CASME Database}
		
	In this scenario, we compare our method with LBP-TOP, completed local binary pattern from three orthogonal planes (CLBP-TOP)~\cite{Pfister2011b}, local ordinary contrast pattern from three orthogonal planes (LOCP-TOP)~\cite{Chan2012}, spatiotemporal local monogenic binary pattern(STLMBP)~\cite{Huang2012}, LBP-SIP~\cite{Wang2014c}, spatiotemporal cuboids descriptor (Cuboids)~\cite{Dollar} and spatiotemporal completed local quantized pattern (STCLQP)~\cite{Huang2016}. Following the parameters setup of~\cite{Huang2016}, we re-implement all comparative methods on CASME database using SVM based on linear kernel, where we divided micro-expression video clip into $8\times 8$ blocks.
	
		\begin{table}[t!]
			\small 
			\captionsetup{justification=centering}
			\caption{Micro-expression recognition accuracies in CASME database. The bold means our proposed methods.}
			\begin{tabular}{|l|c|c|}
				\hline Methods & Block Number &  Accuracy (\%)\\
				\hline LBP-TOP~\cite{Zhao07} & $8\times 8$ & 37.43 \\
				\hline CLBP-TOP~\cite{Pfister2011b} & $8\times 8$ & 45.31 \\ 
				\hline STLMBP~\cite{Huang2012} & $8\times 8$ & 46.20 \\
				\hline LOCP-TOP~\cite{Chan2012} & $8\times 8$ & 31.58\\				
				\hline Cuboids~\cite{Dollar} & - & 33.33 \\
				\hline LBP-SIP~\cite{Wang2014c} & $8\times 8$ & 36.84\\
				\hline STCLQP~\cite{Huang2016} & $8\times 8$ & 57.31\\
				\hline \textbf{STLBP-IIP} & $7\times 3$ & \textbf{59.06} \\
				 \hline \textbf{DiSTLBP-IIP} & $7\times 3$ & \textbf{64.33}\\
				\hline
			\end{tabular}
			\label{tab:CASME}						
			\centering
		\end{table}
	
	Results on recognition accuracy are reported in Table~\ref{tab:CASME}. As seen from the table, LBP-TOP only achieves the accuracy of 37.43\%. LCOP-TOP works worst among all methods. In the comparison algorithms, we see that STCLQP obtains the promising performance on all comparison methods, followed closedly by STLMBP and CLBP-TOP, and more distantly by Cuboids. The reason may be that STCLQP provides more useful information for micro-expression recognition, as STCLQP extracts completed information through sign, magnitude and orientation. But our proposed method STLBP-IIP works better than STCLQP, which is increased by 1.75\%. As well, Laplacian score method further boosts STLBP-IIP, which reaches the best recognition rate of 64.33\% over all methods.

\begin{figure*}[t!]
			\centering
			\includegraphics[width=\linewidth]{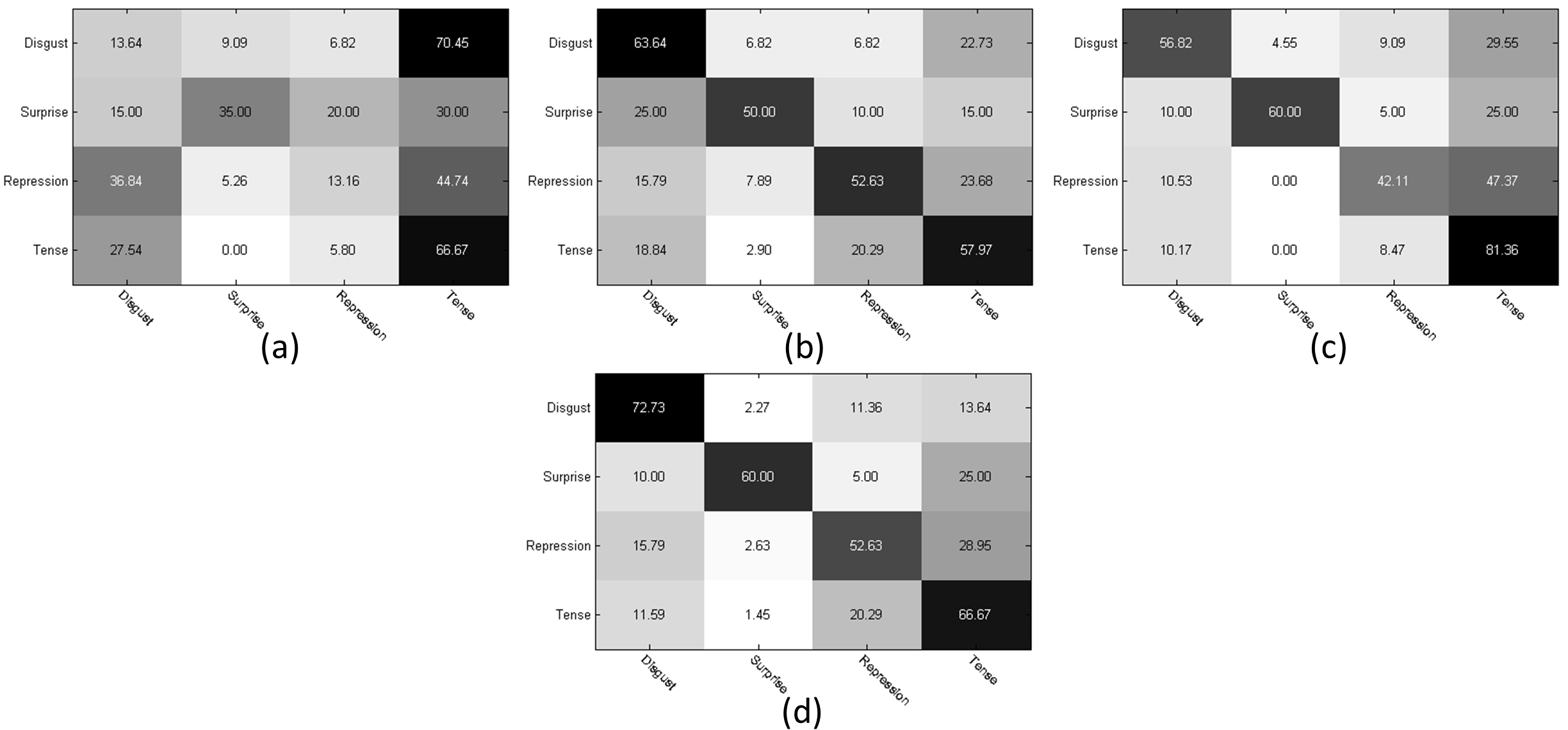}
			\captionsetup{justification=centering}
				\caption{The confusion matrix of (a) LBPTOP, (b) STCLQP, (c) STLBP-IIP, and (d) DiSTLBP-IIP for five micro-expression categorizations on CASME database.}
			\label{fig:CASME_conf}
		\end{figure*}

	The confusion matrix of five micro-expressions is shown in Figure~\ref{fig:CASME_conf}, where we compare our methods with LBP-TOP and STCLQP. It is found that STLBP-IIP performs better on recognizing Surprise and Tense classes, while it works worse than STCLQP on recognizing Disgust and Repression. As seen from Figures~\ref{fig:CASME_conf}(d), recognition performance on Digust and Repression classes is significantly improved by considering discriminative group features for STLBP-IIP. Additionally, we see that Repression and Tense classes are very hard to DiSTLBP-IIP, as they are falsely classified into opposite class. Perhaps it is because Repression and Tense samples are quite similar on CASME database. They are more difficult to be recognized than Disgust and Surprise.
	
	\subsubsection{CASME2 Database}
	
	We compare the recognition rate of our method with the baseline algorithm~\cite{Yan2014}, LBP-TOP~\cite{Zhao07}, LBP-SIP~\cite{Wang2014c}, LOCP-TOP~\cite{Chan2012}. The parameter setup for each method is described as followed:
	
	(1) Following experimental setup of~\cite{Yan2014}, we implement LBP-TOP on $5\times 5$ facial blocks, using radius 3 for LBP operator for three orthogonal planes. For classification, we employ linear-kernel based SVM~\cite{CC01a}. For convenience, we name this method as LBP-TOP+SVM(linear).
		
	(2) We implement the framework of~\cite{Li} based on LBP-TOP~\cite{Zhao07}, LBP-SIP~\cite{Wang2014c} and LOCP-TOP~\cite{Chan2012} as a comparison. Features are extracted on $8\times 8$ facial blocks. According to~\cite{Li}, we firstly use temporal interpolation method~\cite{Zhou2012} to interpolate all videos into 15 frames. Then we implement spatiotemporal features, where the radius and number of neighbors are 3 and 8, respectively. Support Vector Machine (SVM) with Chi-Square Kernel~\cite{CC01a} is used, where the optimal penalty parameter is provided using the three-fold cross validation approach. 

	Comparative performance are presented in Table~\ref{tab:tab1a}. As can be seen, STLBP-IIP is shown to outperform the re-implementation of~\cite{Yan2014}. Its accuracy is increased by 23.88\%. Comparing with LOCP-TOP and LBP-SIP, STLBP-IIP increases the performance by 20.64\% and 22.67\% for micro-expression recognition, respectively. These results demonstrate that STLBP-IIP achieves better performance than LOCP-TOP and LBP-SIP. This is explained by STLBP-IIP preserves the shape for texture descriptor by using the improved integral projection. Comparing with LBP-TOP, DiSTLBP-IIP obtains a significant improvement on micro-expression recognition, since DiSTLBP-IIP extracts the discriminative ability of STLBP-IIP by using feature selection..
	
	Table~\ref{tab:tab2a} shows a comparison to some other dynamic analysis approaches using the recognition rates given in each paper. It should be noted that the results are not directly comparable due to different experimental setups and so forth, but they still give an indication of the discriminative power of each approach. More specifically, leave-one-subject-out cross validation is more strict than leave-sample-out cross validation. Comparing with Table~\ref{tab:tab1a} for LBP-TOP, using leave-one-subject-out cross validation significantly reduces the performance. Our methods outperforms the other methods in almost all cases.	
	
	\begin{table}[t!]
		\small 
		\captionsetup{justification=centering}
		\caption{Comparison under micro-expression recognition rate on CASME2 database. The bold means our proposed methods and * means that we directly extracted the result from their work.}
		\begin{tabular}{|l|c|c|}
			\hline Methods & Block Number & Accuracy (\%)\\
			\hline LBP-TOP+SVM(linear)~\cite{Yan2014} &$5\times 5$ & 38.87\\
			\hline LBP-TOP~\cite{Zhao07} & $8\times 8$ & 39.68 \\ 
			\hline LBP-SIP~\cite{Wang2014c} &$8\times 8$& 40.08\\
			\hline LOCP-TOP~\cite{Chan2012} &$8\times 8$& 42.11 \\
			\hline STCLQP*~\cite{Huang2016} & $8\times 8$ & 58.39 \\
			\hline \textbf{STLBP-IIP} & $6\times 1$& \textbf{62.75}\\
			\hline \textbf{DiSTLBP-IIP} & $6\times 1$& \textbf{64.78}\\
			
			\hline
		\end{tabular}
		\label{tab:tab1a}	
		\centering
	\end{table}
	
	\begin{table*}[t!]
		\small 
		\captionsetup{justification=centering}
		\caption{Performance comparison with the state-of-the-art methods on CASME2 database, where LOSO and LOO represent leave-one-subject-out and leave-one-sample-out cross validation protocols, respectively. The bold means our proposed methods.}
		\begin{tabular}{|l|c|c|c|}
			\hline Methods & Protocol & Task & \hfil Accuracy (\%)\\
			\hline Baseline~\cite{Yan2014} & LOO & Happiness, Surprise, Disgust, Repression and Others & 63.41\\	
			\hline LBP-SIP~\cite{Wang2014c} & LOO & Happiness, Surprise, Disgust, Repression and Others & 67.21\\	
			\hline LSDF~\cite{Wang2014b} & LOO & Happiness, Surprise, Disgust, Repression and Others & 65.44\\
			\hline TICS~\cite{Wang2014} & LOO & Happiness, Surprise, Disgust, Repression and Others & 61.76\\
			
			\hline Monogenic Riesz Wavelet~\cite{Oh2015} & LOSO & Happiness, Surprise, Disgust, Repression and Others & 46.15\\ 
			\hline MDMO~\cite{Liu2016} & LOSO & Positive, Negative, Surprise and Others & 67.37 \\
			\hline STCLQP~\cite{Huang2016} & LOSO & Happiness, Surprise, Disgust, Repression and Others & 58.39 \\
			\hline \textbf{STLBP-IIP} & LOSO & Happiness, Surprise, Disgust, Repression and Others & \textbf{62.75}\\
			\hline \textbf{DiSTLBP-IIP} & LOSO & Happiness, Surprise, Disgust, Repression and Others & \textbf{64.78}\\
			
			\hline
		\end{tabular}
		\label{tab:tab2a}					
		\centering
	\end{table*}
	
\begin{figure*}[t!]
			\centering
			\includegraphics[width=\linewidth]{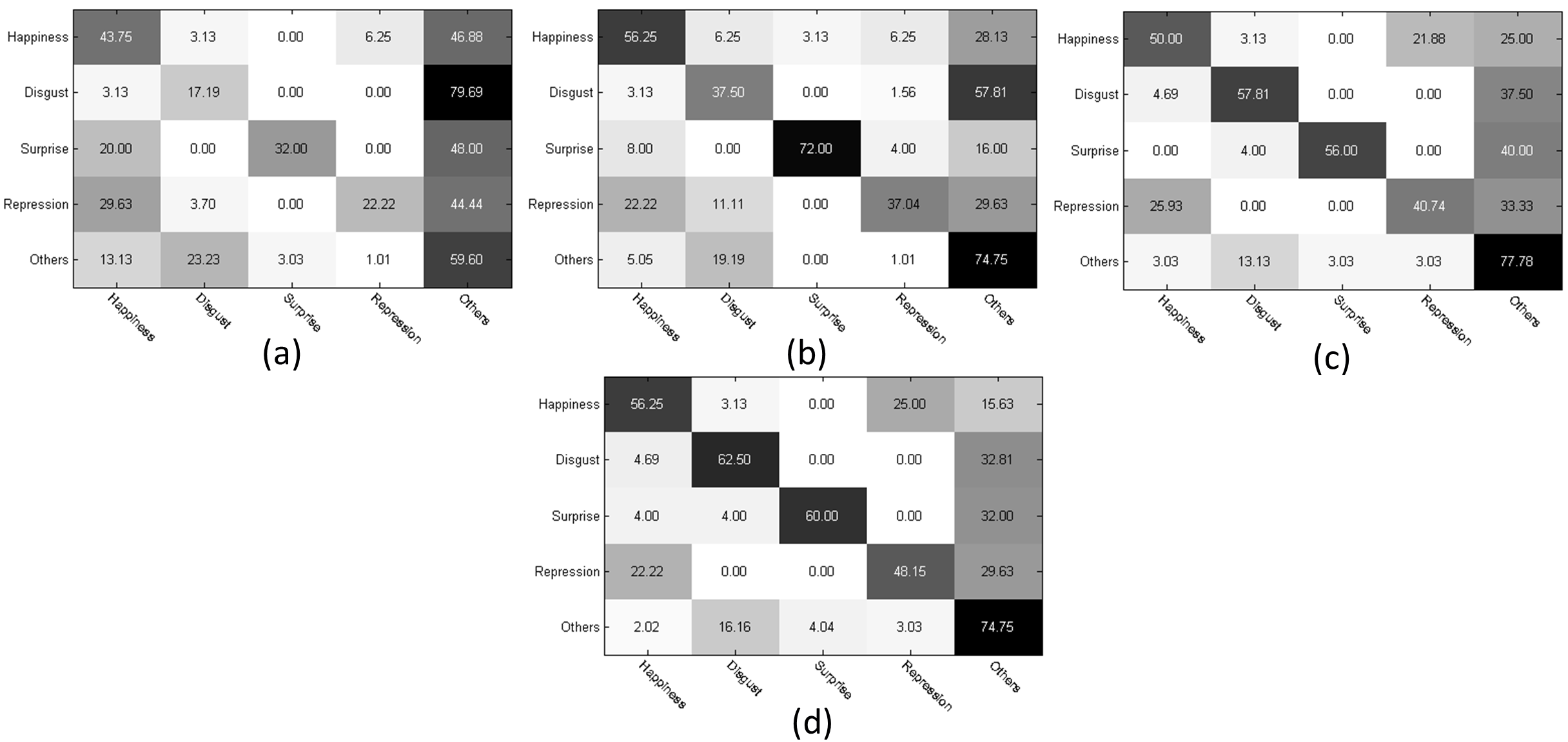}
			\captionsetup{justification=centering}
			\caption{The confusion matrix of (a) LBP-TOP~\cite{Zhao07}, (b) STCLQP~\cite{Huang2016}, (c) STLBP-IIP and (d) DiSTLBP-IIP for five micro-expression categorizations on CASME2 database.}
			\label{fig:CASME2_conf}
		\end{figure*}

	The confusion matrices of six methods is shown in Figure~\ref{fig:CASME2_conf}. As seen from Figures~\ref{fig:CASME2_conf}(c)(d), STLBP-IIP and DisLBP-IIP perform better on recognizing all classes than LBP-TOP. Comparing with STCLQP, STLBP-IIP achieves better performance on three micro-expression classes (Disgust, Repression and Other), while STLCQP outperforms it on recognizing Surprise and Happiness. Our another method, DiSTLBP-IIP outperforms STLCQP at the most of micro-expression classes except Surprise. Unfortunately, DiSTLBP-IIP makes falsely classification of Surprise to Other class. It may be explained that Other class includes some confused micro-expressions similar to Surprise class. From these comparisons, we see that DiSTLBP-IIP has a promising ability to recognize five micro-expressions on CASME2 database, followed by STLBP-IIP and STCLQP.
	
	\subsubsection{SMIC Database}
	
	For SMIC database, we compare our methods with the commonly used spatiotemporal features~\cite{Zhao07,Chan2012, Wang2014c} and feature descriptor based on temporal model~\cite{Jain,Shin}. In our implementation, we used temporal interpolation method (TIM) to normalize each video into 10 frames. As comparison, we use LBP-TOP, LOCP-TOP and LBP-SIP on $5\times 6$ facial blocks for micro-expression recognition. We employ spatiotemporal cuboids feature of~\cite{Dollar} for comparison, where we use k-nearest-neighbor (KNN) classification. We use the same parameter setup to~\cite{Dollar}. Finally, we employ LBP features with conditional random field
	(\textbf{CRF})~\cite{Jain}, geometric features with CRF~\cite{Jain}, dense optical flow with hidden markov model (\textbf{HMM})~\cite{Shin}, for comparison.
		
	The comparison results are reported in Table~\ref{tab:tab1b}. The temporal models with appearance and shape features (LBP+CRF~\cite{Jain}, Shape+CRF~\cite{Jain}, Dense optical flow+HMM~\cite{Shin}) work poorly in micro-expression recognition. Among the temporal model, LBP+CRF~\cite{Jain} gets the best one of 33.54\% for micro-expression recognition. Among all comparative algorithms, STCLQP~\cite{Huang2016} obtains the best recognition rate, followed by LBP-TOP+SVM(linear)~\cite{Li}. Comparing with LBP-TOP~\cite{Zhao07}, STLBP-IIP increases the accuracies of 18.3\% for micro-expression recognition. Comparing with LOCP-TOP~\cite{Chan2012}, the micro-expression recognition performance is increased by 16.47\% for STLBP-IIP. These results demonstrate that STLBP-IIP achieves better performance than geometric features and three spatiotemporal features. Additionally, DiSTLBP-IIP further improves the performance of micro-expression recognition.
	
	\begin{table}[t!]
		\small 
		\captionsetup{justification=centering}
		\caption{Micro-expression recognition accuracies in SMIC database. The bold means our proposed methods and * means that we directly extracted the result from their work.}
		
		\begin{tabular}{|l|c|c|}		
			\hline Methods & Block number & pos/neg/sur (\%)\\
		
			\hline LBP-TOP~\cite{Zhao07} & $5\times 6$ & 42.07 \\
		
			\hline LOCP-TOP~\cite{Chan2012} & $5\times 6$ &  43.90\\
		
			\hline LBP-SIP~\cite{Wang2014c} & $5\times 6$ & 43.29 \\
			
				\hline Cuboids~\cite{Dollar} & - & 37.08 \\
				
				\hline LBP+CRF~\cite{Jain} & - &  33.54 \\
				\hline Shape+CRF~\cite{Jain} & - & 32.93 \\				
				
				\hline Dense optical flow+HMM~\cite{Shin} & - &  20.12 \\
	
			\hline LBP-TOP+SVM(Linear)*~\cite{Li} & $8\times 8$ & 48.78 \\
			\hline STCLQP*~\cite{Huang2016} & $8\times 8$ & 64.02 \\

			\hline \textbf{STLBP-IIP} & $5\times 6$ & \textbf{60.37}\\

			\hline \textbf{DiSTLBP-IIP} & $5\times 6$ & \textbf{63.41}\\

			\hline
		\end{tabular}
		\label{tab:tab1b}		
		\centering
	\end{table}
	
	\section{Conclusion}
	\label{conclusion}
	In the paper, we have shown that the spatiotemporal local binary pattern based on an improved integral projection and its discriminative method achieve the state-of-the-art performance on three facial micro-expression databases. We firstly develop an improved integral projection method to preserve the shape property of micro-expressions and then enhance discrimination of the features for micro-expression recognition. Furthermore, we have presented to use local binary pattern operators to further describe the appearance and motion changes from horizontal and vertical integral projections, well suited for extracting the subtle micro-expressions. Based on Laplacian method, discriminative group features are explored for STLBP-IIP, further enhancing discriminative capability of STLBP-IIP. Experiments on three facial micro-expression databases demonstrate our methods outperform the state-of-the-art methods on micro-expression analysis. In future work, we plan to investigate the gradient and orientation representation, which can be incorporated into our framework.

		
		
		%
		%
		%
		
		\bibliographystyle{natbib.sty}
		\bibliography{IEEEabrv,microegbib}

\begin{thebibliography}{10}
\providecommand{\url}[1]{#1}
\csname url@samestyle\endcsname
\providecommand{\newblock}{\relax}
\providecommand{\bibinfo}[2]{#2}
\providecommand{\BIBentrySTDinterwordspacing}{\spaceskip=0pt\relax}
\providecommand{\BIBentryALTinterwordstretchfactor}{4}
\providecommand{\BIBentryALTinterwordspacing}{\spaceskip=\fontdimen2\font plus
\BIBentryALTinterwordstretchfactor\fontdimen3\font minus
  \fontdimen4\font\relax}
\providecommand{\BIBforeignlanguage}[2]{{%
\expandafter\ifx\csname l@#1\endcsname\relax
\typeout{** WARNING: IEEEtran.bst: No hyphenation pattern has been}%
\typeout{** loaded for the language `#1'. Using the pattern for}%
\typeout{** the default language instead.}%
\else
\language=\csname l@#1\endcsname
\fi
#2}}
\providecommand{\BIBdecl}{\relax}
\BIBdecl

\bibitem{Warren}
G.~Warren, E.~Schertler, and P.~Bull, ``Detecting deception from emotional and
  unemotional cues,'' \emph{Journal of Nonverbal Behavior}, vol.~33, no.~1, pp.
  59--69, 2009.

\bibitem{Frank}
M.~Frank, M.~Herbasz, K.~Sinuk, A.~Keller, and C.~Nolan, ``I see how you feel:
  Training laypeople and professionals to recognize fleeting emotions,'' in
  \emph{International Communication Association}, 2009.

\bibitem{Shreve09}
M.~Shreve, S.~Godavarthy, V.~Manohar, D.~Goldgof, and S.~Sarkar, ``Towards
  macro-and micro-expression spotting in video using strain patterns,'' in
  \emph{Proc. WACV}, 2009, pp. 1--6.

\bibitem{Shreve11}
M.~Shreve, S.~Godavarthy, D.~Goldgof, and S.~Sarkar, ``Macro-and
  micro-exprssion spotting in long videos using spatio-temporal strain,'' in
  \emph{Proc. AFGR}, 2011, pp. 51--56.

\bibitem{Polikovsky}
S.~Polikovsky, Y.~Kameda, and Y.~Ohta, ``Facial micro-expressions recognition
  using high speed camera and 3d-gradient descriptor,'' in \emph{The 3rd
  International Conference on Crime Detection and Preventation}, 2009.

\bibitem{Wu2011}
Q.~Wu, X.~Shen, and X.~Fu, ``The machine knows what you are hiding: an
  automatic micro-expression recognition system,'' in \emph{Affective Computing
  and Intelligence Interaction}, 2011, pp. 152--162.

\bibitem{Ekman1969}
P.~Ekman and W.~Friesen, ``Detecting deception from emotional and unemotional
  cues,'' \emph{Psychiatry}, vol.~32, no.~1, pp. 88--106, 1969.

\bibitem{Yan2013}
W.~Yan, Q.~Wu, Y.~Liu, S.~Wang, and X.~Fu, ``Casme database: A dataset of
  spontaneous micro-expressions collected from neutralized faces,'' in
  \emph{Proc. AFGR}, 2013, pp. 1--7.

\bibitem{Li}
X.~Li, T.~Pfister, X.~Huang, G.~Zhao, and M.~Pietik{\"a}inen, ``A spontaneous
  micro-expression database: Inducement, collection and baseline,'' in
  \emph{Proc. AFGR}, 2013.

\bibitem{Pfister}
T.~Pfister, X.~Li, G.~Zhao, and M.~Pietik{\"a}inen, ``Recognising spontaneous
  facial micro-expressions,'' in \emph{Proc. ICCV}, 2011, pp. 1449--1456.

\bibitem{Yan2014}
W.~Yan, X.~Li, S.~Wang, G.~Zhao, Y.~Liu, Y.~Chen, and X.~Fu, ``{CASME II}: An
  improved spontaneous micro-expression database and the baseline evaluation,''
  \emph{PLOS ONE}, vol.~9, no.~1, pp. 1--8, 2014.

\bibitem{Jung2015}
H.~Jung, S.~Lee, J.~Yim, S.~Park, and J.~Kim, ``Joint fine-tuning in deep
  neural networks for facial expression recognition,'' in \emph{ICCV}, 2015,
  pp. 2983--2991.

\bibitem{Yu2015}
Z.~Yu and C.~Zhang, ``Image based static facial expression recognition with
  multiple deep network learning,'' in \emph{ICMI}, 2015, pp. 435--442.

\bibitem{Ahonen2006}
T.~Ahonen, A.~Hadid, and M.~Pietik\"{a}inen, ``Face description with local
  binary patterns: application to face recognition,'' \emph{IEEE Trans. on
  Pattern Analysis and Machine Intelligence}, vol.~28, no.~12, pp. 2037--2041,
  2006.

\bibitem{Shan09}
C.~Shan, S.~Gong, and P.~W. McOwan, ``Facial expression recognition based on
  local binary pattern: a comprehensive study,'' \emph{Image and Vision
  Computing}, vol.~27, no.~6, pp. 803--816, 2009.

\bibitem{Zhao07}
G.~Zhao and M.~Pietik\"{a}inen, ``Dynamic texture recognition using local
  binary pattern with an application to facial expressions,'' \emph{IEEE Trans.
  on Pattern Analysis and Machine Intelligence}, vol.~29, no.~6, pp. 915--928,
  2009.

\bibitem{Huang2012}
X.~Huang, G.~Zhao, W.~Zheng, and M.~Pietik\"{a}inen, ``Towards a dynamic
  expression recognition system under facial occlusion,'' \emph{Pattern
  Recognition Letters}, vol.~33, no.~16, pp. 2181--2191, 2012.

\bibitem{Davison2014}
A.~Davison, M.~Yap, N.~Costen, K.~Tan, C.~Lansley, and D.~Leightley,
  ``Micro-facial movements: an investigation on spatio-temporal descriptors,''
  in \emph{ECCV workshop on Spontaneous Behavior Analysis}, 2014.

\bibitem{Hernandez}
J.~Ruiz-Hernandez and M.~Pietik{\"a}inen, ``Encoding local binary patterns
  using re-parameterization of the second order gaussian jet,'' in \emph{Proc.
  AFGR}, 2013.

\bibitem{Wang2014}
S.~Wang, W.~Yan, X.~Li, G.~Zhao, and X.~Fu, ``Micro-expression recognition
  using dynamic textures on tensor independent color space,'' in \emph{Proc.
  ICPR}, 2014.

\bibitem{Wang2014b}
S.~Wang, W.~Yan, G.~Zhao, and X.~Fu, ``Micro-expression recognition using
  robust principal component analysis and local spatiotemporal directional
  features,'' in \emph{ECCV workshop on Spontaneous Behavior Analysis}, 2014.

\bibitem{Wang2014c}
Y.~Wang, J.~See, R.~Phan, and Y.~Oh, ``{LBP} with six interaction points:
  Reducing redundant information in {LBP-TOP} for micro-expression
  recognition,'' in \emph{Proc. ACCV}, 2014.

\bibitem{Guo2016}
Y.~Guo, C.~Xue, Y.~Wang, and M.~Yu, ``Mirco-expression recognition based on
  cbp-top feature with elm,'' \emph{International Journal for Light and Electro
  Optics}, vol. 127, no.~4, p. 2404, 2009.

\bibitem{Oh2015}
X.~He, D.~Cai, and P.~Niyogi, ``Monogenic riesz wavelet representation for
  micro-expression recognition,'' in \emph{IEEE International Conference on
  Digital Signal Processing}, 2015, pp. 1237--1241.

\bibitem{Huang2016}
X.~Huang, G.~Zhao, X.~Hong, W.~Zheng, and M.~Pietik{\"a}inen, ``Spontaneous
  facial micro-expression analysis using spatiotemporal completed local
  quantized patterns,'' \emph{Neurocomputing}, no. 175, pp. 564--578, 2016.

\bibitem{Kotsia2008}
I.~Kotsia, S.~Zafeiriou, and I.~Pitas, ``Texture and shape information fusion
  for facial expression and facial action unit recognition,'' \emph{Pattern
  Recognition}, vol.~41, no.~3, pp. 833--851, 2008.

\bibitem{Houam}
L.~Houam, A.~Hafiane, A.~Boukrouche, E.~Lespessailles, and R.~Jennane, ``One
  dimensional local binary pattern for bone texture characterization,''
  \emph{Pattern Analysis and Applications}, vol.~17, pp. 179--193, 2014.

\bibitem{Benzaoui2013}
A.~Benzaoui and A.~Boukrouche, ``Face recognition using 1{DLBP} texture
  analysis,'' in \emph{The International Conference on Future Computational
  Technologies and Applications}, 2013, pp. 14--19.

\bibitem{Benzaoui2015}
A.~Benzaoui, ``Face recognition using 1{DLBP}, {DWT} and {SVM},'' in \emph{The
  International Conference on Control, Engineering and Information Technology},
  2015, pp. 1--6.

\bibitem{Robinson}
D.~Robinson and P.~Milanfar, ``Fast local and global projection-based methods
  for affine motion estimation,'' \emph{Journal of Mathematical Imaging and
  Vision}, vol.~18, pp. 35--54, 2003.

\bibitem{Mateos03}
G.~Mateos, ``Refining face tracking with integral projection,'' in \emph{Proc.
  AVBPA}, 2003, pp. 360--368.

\bibitem{Mateos07}
G.~Mateos, A.~Ruiz-Garcia, and P.~Lopez-de Teruel, ``Human face processing with
  1.5d model,'' in \emph{Proc. AMFG}, 2007, pp. 220--234.

\bibitem{Carcia2002}
G.~Carcia-Mateos, A.~Ruiz, and P.~L. de~Teruel, ``Face detection using integral
  projection models,'' in \emph{Joint IAPR International Workshops SSPR 2002
  and SPR 2002 Windsor}, 2002, pp. 644--653.

\bibitem{Zhao2009}
G.~Zhao and M.~Pietik{\"a}inen, ``Boosted multi-resolution spatiotemporal
  descriptors for facial expression recognition,'' \emph{Pattern Recognition
  Letters}, vol.~30, no.~12, pp. 1117--1127, 2009.

\bibitem{He2005}
X.~He, D.~Cai, and P.~Niyogi, ``Laplacian score for feature selection,'' in
  \emph{Proc. NIPS}, 2005.

\bibitem{Benzaoui}
A.~Benzaoui and A.~Boukrouche, ``Face recognition using 1dlbp texture
  analysis,'' in \emph{Proc. FCTA}, 2013, pp. 14--19.

\bibitem{Wright2009}
J.~Wright, A.~Ganesh, S.~Rao, Y.~Peng, and Y.~Ma, ``Robust principal component
  analysis: exact recovery of corrupted low-rank matrices via context
  optimization,'' in \emph{Proc. NIPS}, 2009, pp. 2080--2088.

\bibitem{Ojala02}
T.~Ojala, M.~Pietik\"{a}inen, and T.~M{\"a}enp{\"a}{\"a}, ``Multiresolution
  gray-scale and rotation invariant texutre classification with local binary
  pattern,'' \emph{IEEE Trans. on Pattern Analysis and Machine Intelligence},
  vol.~24, no.~7, pp. 971--987, 2002.

\bibitem{CC01a}
C.~Chang and C.~Lin, ``{LIBSVM}: A library for support vector machines,''
  \emph{ACM Transactions on Intelligent Systems and Technology}, vol.~2, pp.
  27:1--27:27, 2011.

\bibitem{Huang2015}
X.~Huang, S.-J. Wang, G.~Zhao, and M.~Pietik{\"a}ine, ``Facial micro-expression
  recognition using spatiotemporal local binary pattern with integral
  projection,'' in \emph{Proc. ICCV Workshop}, 2015, pp. 1--9.

\bibitem{Pfister2011b}
T.~Pfister, X.~Li, G.~Zhao, and M.~Pietik{\"a}inen, ``Differentiating
  spontaneous from posed facial expressions within a generic facial expression
  recognition framework,'' in \emph{Proc. ICCV}, 2011, pp. 868--875.

\bibitem{Chan2012}
C.~Chan, B.~Goswami, J.~Kittler, and W.~Christmas, ``Local ordinal contrast
  pattern histograms for spatiotemporal, lip-based speaker authentication,''
  \emph{IEEE Trans. on Information Forensics and Security}, vol.~2, no.~7, pp.
  602--612, 2012.

\bibitem{Dollar}
P.~Dollar, V.~Rabaud, G.~Cottrell, and S.~Belongie, ``Behavior recognition via
  sparse spatio-temporal features,'' in \emph{Proc. VSPETS}, 2005, pp. 65--72.

\bibitem{Zhou2012}
Z.~Zhou, G.~Zhao, Y.~Guo, and M.~Pietik{\"a}inen, ``An image-based visual
  speech animation system,'' \emph{IEEE Trans. on Circuits and Systems for
  Video Technology}, vol.~22, no.~10, pp. 1420--1432, 2012.

\bibitem{Liu2016}
Y.~Liu, J.~Zhang, W.~Yan, S.~Wang, G.~Zhao, and X.~Fu, ``A main directional
  mean optical flow feature for spontaneous micro-expression recognition,''
  \emph{IEEE Trans. on Affective Computing}, no.~99, p.~1, 2016.

\bibitem{Jain}
S.~Jain, C.~Hu, and J.~Aggarwal, ``Facial expression recognition with temporal
  modeling of shapes,'' in \emph{Proc. ICCV}, 2011, pp. 1642--1649.

\bibitem{Shin}
G.~Shin and J.~Chun, ``Spatio-temporal facial expression recognition using
  optical flow and hmm,'' \emph{Software Engineering, Artificial Intelligence,
  Network, and Parallel/Distributed Computing}, pp. 27--38, 2008.

\end{thebibliography}
		
		%

		
		

	\end{document}